\documentclass[runningheads]{llncs}
\usepackage[dvipdfmx]{graphicx}
\usepackage[dvipdfmx]{color}
\usepackage{comment}
\usepackage{amsmath,amssymb} 
\usepackage{color}
\usepackage{xcolor}
\usepackage{enumitem}
\usepackage{hhline}
\usepackage{threeparttable}
\usepackage{multirow}
\usepackage{mathtools}
\usepackage{setspace}
\usepackage{array}
\newcolumntype{L}[1]{>{\raggedright\let\newline\\\arraybackslash\hspace{0pt}}m{#1}}
\newcolumntype{C}[1]{>{\centering\let\newline\\\arraybackslash\hspace{0pt}}m{#1}}
\newcolumntype{R}[1]{>{\raggedleft\let\newline\\\arraybackslash\hspace{0pt}}m{#1}}

\makeatletter
\newcommand{\thickhline}{%
    \noalign {\ifnum 0=`}\fi \hrule height 1pt
    \futurelet \reserved@a \@xhline
}
\newcolumntype{"}{@{\hskip\tabcolsep\vrule width 1pt\hskip\tabcolsep}}
\makeatother
\graphicspath{ {./figures/} }

\begin{document}
\pagestyle{headings}
\mainmatter
\def\ECCVSubNumber{3947}  

\newcommand{\argmax}{\mathop{\rm arg~max}\limits}
\newcommand{\argmin}{\mathop{\rm arg~min}\limits}

\makeatletter
\def\@fnsymbol#1{\ensuremath{\ifcase#1\or *\or \dagger\or \ddagger\or
   \mathsection\or \mathparagraph\or \|\or **\or \dagger\dagger
   \or \ddagger\ddagger \else\@ctrerr\fi}
}
\makeatother

\title{Privacy Preserving Visual SLAM}
\titlerunning{Privacy Preserving Visual SLAM}

\author{Mikiya Shibuya\thanks{The authors assert equal contribution and joint first authorship.}\inst{1,2} \and
Shinya Sumikura\inst{*\ 1} \and
Ken Sakurada\inst{*\ 1}}
\authorrunning{M. Shibuya, S. Sumikura, K. Sakurada}
\institute{National Institute of Advanced Industrial Science and Technology (AIST) \and
Tokyo Institute of Technology \\
\email{\{mikiya-shibuya,sumikura.shinya,k.sakurada\}@aist.go.jp} \\
\email{shibuya@m.titech.ac.jp}
}
\maketitle

\begin{abstract}
This study proposes a privacy-preserving Visual SLAM framework for estimating camera poses and performing bundle adjustment with mixed line and point clouds in real time. 
Previous studies have proposed localization methods to estimate a camera pose using a line-cloud map for a single image or a reconstructed point cloud.
These methods offer a scene privacy protection against the inversion attacks by converting a point cloud to a line cloud, which reconstruct the scene images from the point cloud. 
However, they are not directly applicable to a video sequence because they do not address computational efficiency. 
This is a critical issue to solve for estimating camera poses and performing bundle adjustment with mixed line and point clouds in real time. 
Moreover, there has been no study on a method to optimize a line-cloud map of a server with a point cloud reconstructed from a client video because any observation points on the image coordinates are not available to prevent the inversion attacks, namely the reversibility of the 3D lines. 
The experimental results with synthetic and real data show that our Visual SLAM framework achieves the intended privacy-preserving formation and real-time performance using a line-cloud map.

\keywords{Visual SLAM, privacy, line cloud, point cloud}
\end{abstract}

\section{Introduction}
\label{sec:introduction}
Localization and mapping from images are fundamental problems in the field of computer vision. They have been exhaustively studied for robotics and augmented/mixed reality (AR/MR) \cite{davison2008monoslam,klein2007georg,sattler2011fast}.
These applications are divided into three main types, where the 6 degree-of-freedom (DOF) camera pose is:
(i) in unmeasured regions to be  estimated simultaneously with the 3D map through either Structure from Motion (SfM) \cite{agarwal2011building,cui2017hsfm} or Visual SLAM~\cite{engel2018direct,engel2014lsdslam,murartal2015orbslam};
(ii) in measured regions to be estimated by solving 2D--3D matching between the image and the 3D map 
and (iii) in both of measured and unmeasured regions, a camera passes through the entire regions. 
Because of the complexity of this field in computer vision, this study focuses on the literature regarding the applications of (ii) and (iii).

Recent studies have revealed a risk of privacy preservation that 3D points and their descriptors can be inverted to synthesize the original scene images \cite{pittaluga2019revealing}. 
To prevent this privacy risk, Speciale \textit{et al.} proposed a privacy-preserving method which converts a 3D point cloud to a 3D line cloud to make the inversion attack difficult \cite{speciale2019privacylocalization,speciale2019privacyqueries}.
However, in the case of camera pose estimation of a single image, the problem after the conversion changes from three 2D point--3D point correspondences (p3P) 
\cite{fischler1981random,haralick1991analysis,quan1999linear} to six 2D point--3D line correspondences (p6L) \cite{speciale2019privacylocalization}, 
which causes the amount of computation to increase and the accuracy to deteriorate.
Moreover, for the corresponding search with 2D points, the computational cost and the matching error ratio for a 3D line are higher than those for a 3D point. 
Hence, it is difficult to directly apply the localization method with p6L to a real-time application with a video sequence, such as Visual SLAM.

\begin{figure*}[t]
\begin{center}
\vspace{-2mm}
\includegraphics[width=12cm,bb=0 30 1946 336]{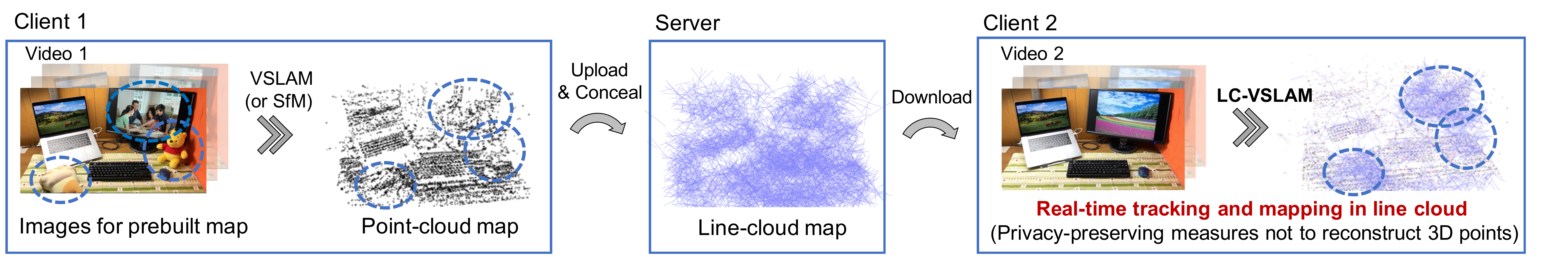}
\caption{
Example of LC-VSLAM application.
}
\label{fig:abst_lcvslam}
\vspace{-5mm}
\end{center}
\end{figure*}

For SfM and Visual SLAM, bundle adjustment (BA) is utilized to optimize camera pose and 3D points \cite{triggs1999bundle,wu2011multicore}. 
In a standard BA, the parameters are optimized by minimizing the error function with distances between the reprojected points and the corresponding 2D points. 
However, there are two new problems for BA with regard to a line-cloud map from a server. 
First, BA for the line-cloud map demands an additional definition of every new error function between a 2D point on a client image and the corresponding 3D line from a server. 
Second, to ensure the irreversibility of a line cloud to the original point cloud, it is inevitable to integrate the line cloud with 
the point cloud and to globally optimize them without the 2D point coordinates on the keyframes of the line cloud.

To overcome these difficulties, we propose a Visual SLAM framework for real-time relocalization, tracking, and BA with a map mixed with lines and points (Fig. \ref{fig:abst_lcvslam} and \ref{fig:overview_lcvslam}), which we call {\it Line-Cloud Visual SLAM} (LC-VSLAM). The main contributions of this study are three-fold.
\renewcommand{\labelitemi}{$\bullet$}
\begin{itemize}
 \item Efficient relocalization and tracking with 3D points reconstructed by Visual SLAM of a client.
 \item Motion-only, rigid-stereo, local, and global bundle adjustments for mixed line and point clouds.
 \item Creation of unified framework for various types of projection models, such as perspective, fisheye, and equirectangular.
\end{itemize}

First, matching between local 3D points reconstructed with Visual SLAM by a client and a line cloud enables fast and accurate relocalization (Sec. \ref{subsec:relocalization_loopdetection}). 
Moreover, discretizing the 3D line to 3D points speeds up the 2D--3D matching to achieve real-time tracking (Sec. \ref{subsec:matching}).

Second, we propose four types of bundle adjustments for mixed line and point clouds, motion-only, rigid-stereo, local, and global BAs, 
depending on the optimization parameters. 
The 3D lines are simultaneously optimized with the camera poses and 3D points by defining the covariance of the 3D line with that of the original 3D point, whose value in the direction of the line is infinite.
The covariance is used to calculate the reprojection error between the 3D line and the corresponding 2D point (or line). 
In the global BA, a whole map which has already included a line cloud from a server is optimized by adding the virtual observations of 3D lines on the line-cloud keyframes (Sec. \ref{subsec:line_bundle_adjustments}).

Finally, we propose a unified framework that can be applied to various types of projection models by reason of the matching efficiency, 
where 3D lines are discretized to 3D points (Sec. \ref{subsec:matching}).
The reprojection error between the 3D line and the virtual observation is defined as the difference between the normal vectors of the planes consisting of the lines and the origin of the local camera coordinates (Sec. \ref{subsec:line_bundle_adjustments}).

In Section \ref{sec:related_work}, we summarize the related work. 
In Section \ref{sec:proposed_method}, we explain the details of the proposed framework. 
In Section \ref{sec:experiments}, we present the experimental results. 
Finally, in Section \ref{sec:conclusion}, we present our conclusions.

\section{Related Works}
\label{sec:related_work}
\subsection{Visual SLAM}
\label{subsec:related_work_visual_slam}
Visual SLAM is broadly utilized for environment mapping, localization in robotics, and camera tracking frameworks in AR/MR applications.
The Visual SLAM algorithms are generally divided into three kinds of methods: feature-based~\cite{davison2008monoslam,klein2007georg,murartal2015orbslam,murartal2017orbslam2}, direct~\cite{engel2018direct,engel2014lsdslam,newcombe2011dtam}, and learning-based~\cite{detone2017superpoint,tang2019gcnv2,tateno2017cnnslam,yang2018dvso,yi2016lift,zhou2018deeptam}.
The feature-based methods pertain to camera tracking and scene mapping with feature points extracted from images~\cite{alcantarilla2012kaze,bay2997surf,lowe2004distinctive,rublee2011orb}.
The direct methods, in contrast, focus on minimization of photometric errors indicating the difference of the intensity between two frames.

Recently, a combination of Convolutional Neural Networks (CNNs) and either of the aforementioned kinds of algorithms (feature-based or direct) has been under extensive investigation.
The feature-based methods use CNN-based architectures in conventional algorithms to detect and describe their feature points~\cite{detone2017superpoint,tang2019gcnv2,yi2016lift}.
For the direct methods, CNN-based depth prediction techniques are utilized for the initialization of depth estimation~\cite{tateno2017cnnslam,yang2018dvso}.
As opposed to the fusion of conventional Visual SLAM and learning-based methods, end-to-end tracking and mapping methods based on Deep Neural Networks (DNNs) have been recently studied~\cite{zhou2018deeptam}.

The feature-based methods can localize frames in a prebuilt map quickly and accurately~\cite{galvezlopez2012bagofbinary,dominik2018hbst}.
These characteristics are required for our LC-VSLAM to localize camera pose against a prebuilt 3D line cloud, to track camera trajectory, and to simultaneously expand the map.
Therefore, we constructed the LC-VSLAM algorithm based on the feature-based Visual SLAM.

\subsection{Map Representation with Line Cloud}
In conventional AR/MR applications, each client downloads a prebuilt map created by other clients from a server and performs localization/tracking based on the map. In this case, the clients share only the 3D point cloud and its optional attributes (e.g., color, descriptor, and visibility of each point and camera poses).
However, Pittaluga \textit{et al.} proved that fine images at arbitrary viewpoints can be restored only with the sparse point cloud and its optional attributes~\cite{pittaluga2019revealing}.
They referred to this restoration as an inversion attack.

To address this problem, Speciale \textit{et al.} proposed a map representation based on a 3D line cloud~\cite{speciale2019privacylocalization}.
They also formulated a method to localize an image in the prebuilt line cloud.
The line cloud is built by converting each 3D point to a 3D line that has a random orientation and passes through the original point.
It is quite difficult to directly restore the original point cloud from the line cloud because the point coordinates can be reparameterized arbitrarily on the corresponding line.

To the best of our knowledge, there has been no study on how to track camera poses continuously in real time with a 3D line cloud.
As a straightforward method, the camera pose of every frame can be estimated successively using the p6L solver proposed in \cite{speciale2019privacylocalization}.
However, the p6L solver has a much larger computational cost than the typical p3P solvers~\cite{fischler1981random,haralick1991analysis,quan1999linear}.
In contrast, some methods achieve Visual SLAM using edges in a scene as landmarks like feature points~\cite{dong2017linebased,huizhong2015structslam,puimarola2017plslam}.
These methods build 3D lines based on the structure and color distribution in a scene.
That is, the 3D lines explicitly represent the scene structures. 
Our method, however, to be resistant against inversion attacks, avoids such explicitness by utilizing the randomly oriented 3D line cloud.

\subsection{Bundle Adjustment for Map Optimization}
Conventional Visual SLAM and SfM methods largely utilize pose graph optimization (PGO)~\cite{grisetti2010pgo,kummerle2011g2o,strasdat2010scale} and bundle adjustment (BA)~\cite{lourakis2009sba,triggs1999bundle,wu2011multicore} for accurate pose estimation and map construction.
PGO can be almost directly applied to loop closure for a line cloud, but the conventional BA cannot.
This is because a reprojection error that constrains a 3D line and a 2D feature point has not been defined.
For the Visual SLAM methods based on structural edges, point-to-line distances between a 2D line and two endpoints of a reprojected 3D line segment are used as a reprojection error between 2D and 3D lines~\cite{dong2017linebased,puimarola2017plslam}.
However, this formulation targets the structure-based edges, i.e., the 3D lines which explicitly represent the scene structures; thus, they cannot be applied to tracking and mapping with a prebuilt map of randomly oriented 3D lines.

We therefore propose a reprojection constraint between randomly oriented 3D lines and 2D feature points in order to conduct BA for the map representation with mixed line and point clouds.
In its formulation, an error ellipse of each 3D line is decided according to the covariance of the corresponding 3D point before being converted to the 3D line.
The error ellipse of the 3D line has not been considered in the previous study~\cite{speciale2019privacylocalization}.
Furthermore, the proposed algorithms do not depend on the difference of projection models.
Hence, the real-time LC-VSLAM framework can be realized in various types of projection models.

\section{Proposed Method}
\label{sec:proposed_method}

\begin{figure*}[t]
\begin{center}
\includegraphics[width=12cm]{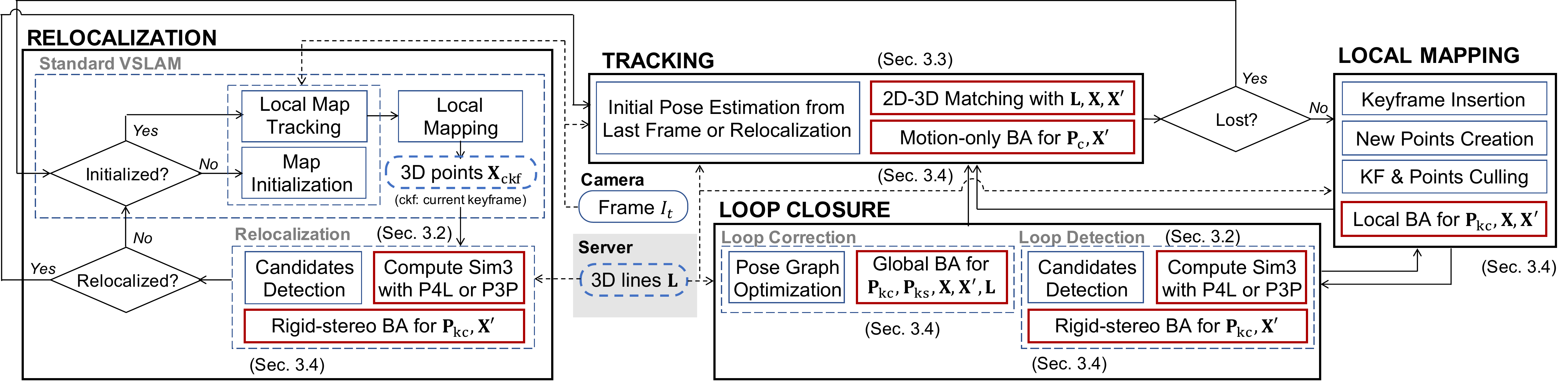}
\vspace{0mm}
\caption{
Overview of LC-VSLAM system. It should be noted here that the three threads run in parallel: tracking, local mapping, and loop closure.
}
\label{fig:overview_lcvslam}
\vspace{-5mm}
\end{center}
\end{figure*}

\subsection{System Overview}
\label{subsec:system_overview}
The proposed LC-VSLAM system consists of four modules: relocalization, tracking, local mapping, and loop closure (Fig. \ref{fig:overview_lcvslam}). 
The system works to estimate the parameters set for the following camera poses and 3D mapping: 
\begin{quote}
\begin{enumerate}[label={(\arabic*)}]
  \item Camera pose of the current client frame $\mathbf{P}_\text{c}$,
  \item Camera poses of client keyframes $\mathbf{P}_\text{kc}$,
  \item Camera poses of server keyframes for 3D lines $\mathbf{P}_\text{ks}$,
  \item 3D lines $\mathbf{L}$,
  \item 3D points reconstructed only from 2D points $\mathbf{X}$,
  \item 3D points reconstructed from 3D lines and 2D points $\mathbf{X'}$.
\end{enumerate}
\end{quote}

First, for relocalization in a line cloud, the system performs a standard Visual SLAM \cite{murartal2015orbslam,zhou2018deeptam} for video sequence input $I_{1:t}$ to reconstruct the local 3D points of the current keyframe $\mathbf{X}_\text{ckf}$  (Sec. \ref{subsec:relocalization_loopdetection}). 
The camera poses of the keyframes in the line cloud $\mathbf{P}_\text{kc}$ are calculated with $\mathbf{X}_\text{ckf}$ by computing Sim(3) 
with four 3D point–3D line (P4L) \cite{sweeney2014gdls} or three 3D point–3D point (P3P) \cite{eggert1997estimating,horn1987closed,umeyama1991least} correspondences after the candidate detection based on DBOW \cite{galvezlopez2012bagofbinary}.  
Then, the camera poses $\mathbf{P}_\mathrm{kc}$ and the reconstructed 3D points $\mathbf{X}'$ are optimized in the 
rigid-stereo bundle adjustment (Sec. \ref{subsec:line_bundle_adjustments}). 
The loop detection performs a similar processing operation (Sec. \ref{subsec:relocalization_loopdetection}). 
After the relocalization in the line cloud, the other three LC-VSLAM modules (tracking, local mapping, and loop closure) start.

The tracking module continuously estimates the camera pose for the current frame. 
The tentative camera pose is estimated by assuming a linear motion of the camera. 
In the 2D point–3D line matching, 3D lines are discretized to 3D points to improve the computational efficiency (Sec. \ref{subsec:matching}).
Using all the correspondences of the 2D point–3D point and the 2D point–3D line, the motion-only bundle adjustment optimizes the camera pose of the current frame $\mathbf{P}_\text{c}$ and the reconstructed 3D points $\mathbf{X'}$ (Sec. \ref{subsubsec:motion-only-ba}).

In the local mapping module, 3D points $\mathbf{X}, \mathbf{X'}$ are newly created or restored using the keyframes with the camera pose estimated in the tracking module, according to the 2D point–2D point and the 2D point–3D line correspondences. 
Subsequently, the local bundle adjustment optimizes the camera poses of the client keyframes $\mathbf{P}_\text{kc}$ and the reconstructed 3D points $\mathbf{X},\mathbf{X'}$ simultaneously (Sec. \ref{subsubsec:local-ba}).

For correcting errors of the 3D lines and points, the loop-closure module detects the loops in the same manner as relocalization. 
After the pose graph optimization \cite{grisetti2010pgo,strasdat2010scale}, the global bundle adjustment optimizes all of the parameters for the map $\mathbf{P}_\text{kc},\mathbf{P}_\text{ks},\mathbf{X},\mathbf{X'},\mathbf{L}$ by introducing the virtual observations of the 3D lines on the line-cloud keyframes (Sec. \ref{subsubsec:global-ba}).

\subsection{Relocalization and Loop Detection with a Line Cloud}
\label{subsec:relocalization_loopdetection}

\begin{figure}[!t]
\hfil
$\begin{array}{c|c}\scriptsize
   \includegraphics[width=6cm]{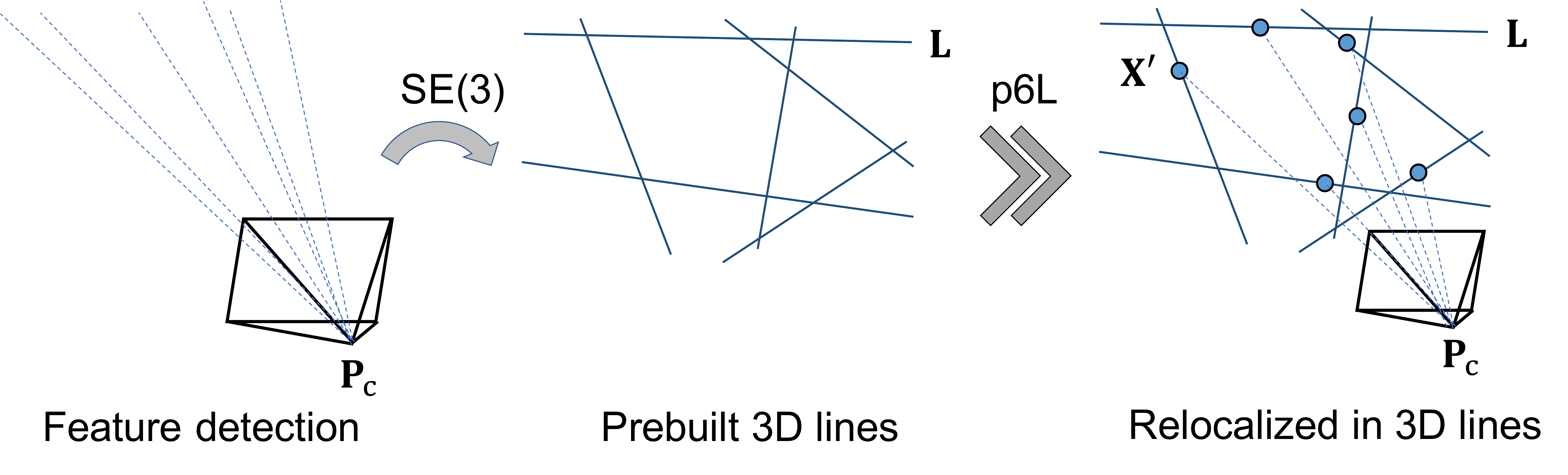}
&
   \includegraphics[width=6cm]{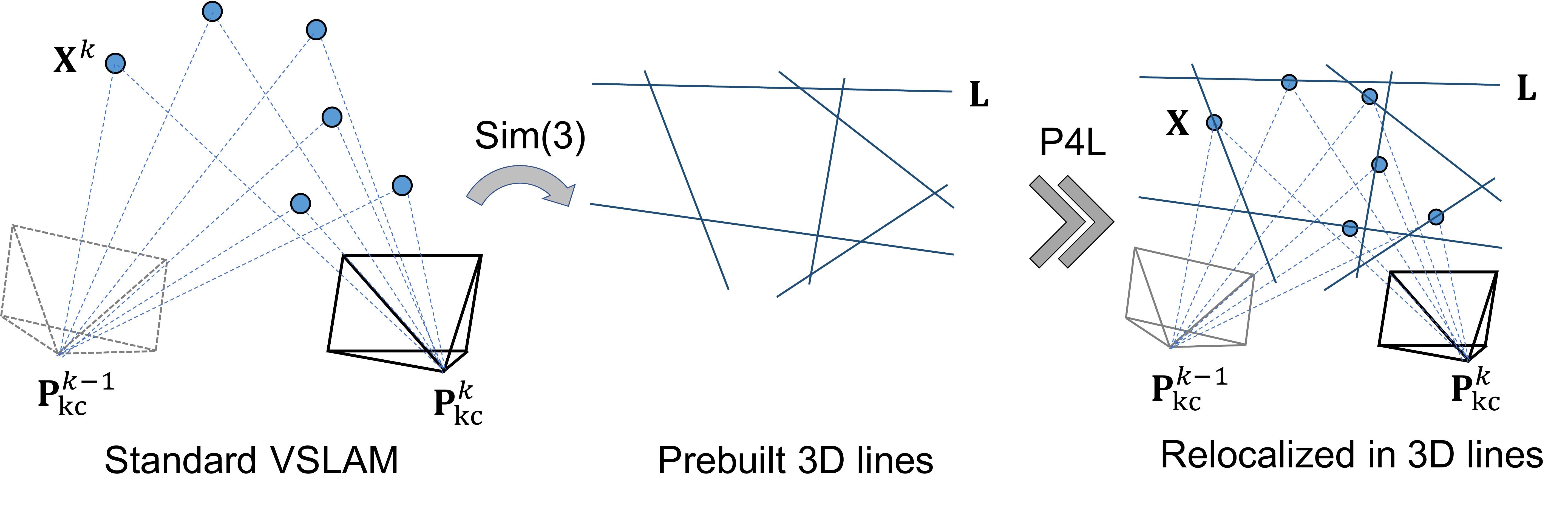} \\
\text{\scriptsize (a) Single shot} & \text{\scriptsize (b) Video sequence (ours)} \\
\end{array}$
  \caption{Overview of relocalization and loop detection with a line cloud.}
  \label{fig:overview_relocalization}
\end{figure}

In this study, we assume that the visibility of 3D lines $\mathbf{L}$ from a server for the keyframes $\mathbf{P}_\mathrm{ks}$ is known.
Hence, for the global localization problem using a line-cloud, we utilize a bag-of-words strategy such as DBOW \cite{galvezlopez2012bagofbinary}, in the same manner as in a standard Visual SLAM \cite{murartal2015orbslam,zhou2018deeptam} to efficiently detect loop candidates.
After the loop candidate detection, the geometric verification with a RANSAC-based solver rejects the outliers of their descriptor matches
\cite{RANSAC}.
As shown in Figure \ref{fig:overview_relocalization}(a), the increase in the computational cost of the p6L solver \cite{speciale2019privacylocalization}, compared to that of the typical p3P solvers, prevents real-time processing due to requiring more points to solve a minimal problem \cite{fischler1981random,haralick1991analysis,quan1999linear}.
Therefore, we utilize local 3D points of the current keyframe $\mathbf{X}_\mathrm{ckf}$, which are reconstructed by a standard Visual SLAM of a client, to match with the 3D lines $\mathbf{L}$ [Fig. \ref{fig:overview_relocalization}(b)]. 
More concretely, we utilize four 3D point-3D line correspondences (P4L) to calculate the relative Sim(3) pose $\Delta \mathbf{P}_\mathrm{kc}^\mathrm{Sim3}$.
The P4L solver \cite{sweeney2014gdls} is more efficient than the p6L one. (The p6L solver cannot be directly applied to the Sim(3) estimation for the scale drift-aware loop closure \cite{strasdat2010scale}.)

In cases where 3D points have already been reconstructed in the line-cloud map (e.g., relocalization after tracking loss and loop detection after exploring the line cloud), both 3D lines and points are utilized for the Sim(3) estimation.
To be more precise, we use P4L if $N_\mathrm{PL} / N_\mathrm{PP} > 4/3$ and P3P otherwise, where $N_\mathrm{PL}$ and $N_\mathrm{PP}$ represent the numbers of 3D point-3D line and the 3D point-3D point correspondences, respectively.
After the initial estimation, the pose graph optimization is conducted with the relative camera pose \cite{grisetti2010pgo,strasdat2010scale}, and the rigid-stereo BA optimizes the camera pose $\mathbf{P}_\mathrm{kc}^\mathrm{ckf}$ and the reconstructed 3D points $\mathrm{X'}$ (Sec. \ref{subsec:line_bundle_adjustments}).

\subsection{2D--3D Matching with 3D Lines and Points}
\label{subsec:matching}
\begin{figure}[!t]
\vspace{2mm}
\hfil
$\begin{array}{cc}\scriptsize
 {\includegraphics[width=3.0cm,bb=0 0 347 261]{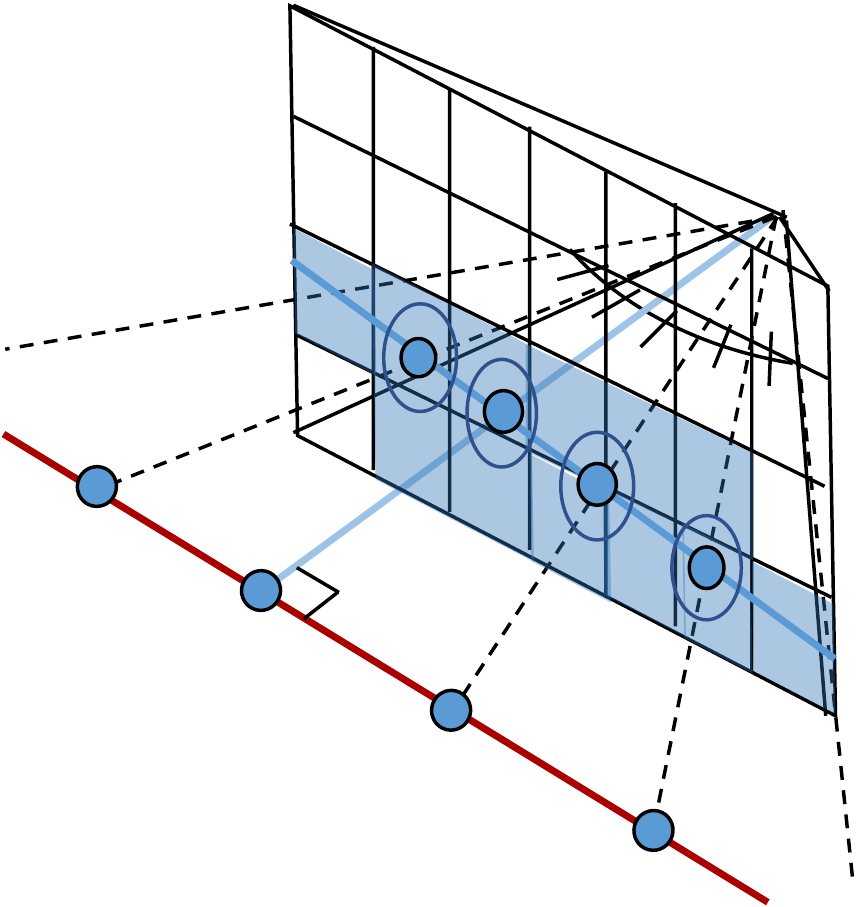} \hspace{2mm} }&
   { \hspace{2mm} \includegraphics[width=7cm,bb=-100 0 823 286]{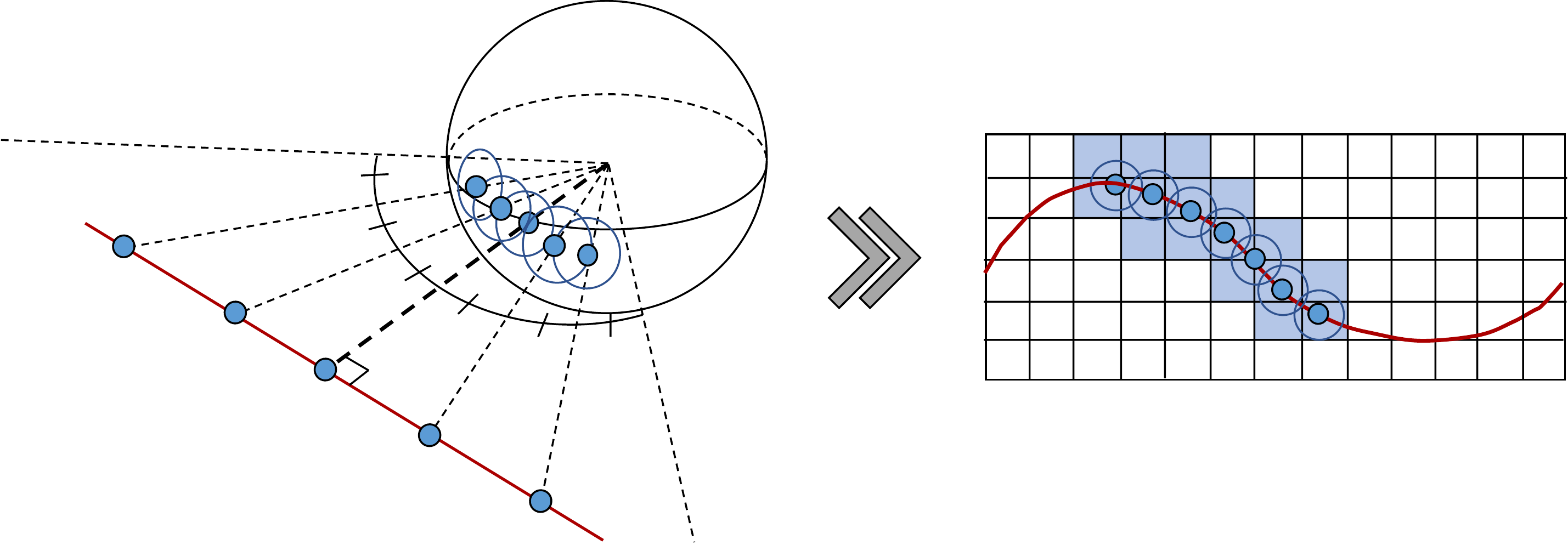}}\\
\text{\scriptsize (a) Perspective projection model \hspace{10mm}} & \text{\scriptsize \hspace{10mm} (b) Equirectangular projection model} \\
\end{array}$
  \vspace{0mm}
  \caption{Overview of matching between 2D points and discretized 3D lines.}
  \label{fig:overview_matching}
  \vspace{0mm}
\end{figure}

Real-time tracking with a line cloud requires a fast search between corresponding 2D points and 3D lines as well as between 2D and 3D points for standard feature-based methods \cite{murartal2015orbslam}.
However, especially for the equirectangular projection model, efficient search is difficult because the reprojected 3D lines are not straight to correspond the 3D points in the image coordinates.
Hence, in our system, a 3D line is discretized to 3D points, and they are reprojected onto the image.
This discretization strategy brings about an advantage of efficient search for corresponding 2D points that narrows search ranges of distances and image coordinates.
Moreover, this method can be directly applied to various types of projection models, such as the perspective and equirectangular models.
Figure \ref{fig:overview_matching} shows the overview of this 2D point-3D line matching.

\subsection{Bundle Adjustments with a Line Cloud}
\label{subsec:line_bundle_adjustments}
To achieve bundle adjustments with a line cloud, first we define the information matrix of a 3D line with the covariance matrix of the original 3D point.
Next, we also define error metrics between a 2D point (or line) and a 3D line.
Finally, utilizing the error metrics, we introduce new error functions for each bundle adjustment.

\begin{figure}[!t]
\hfil
\includegraphics[width=12cm,bb=0 0 1670 409]{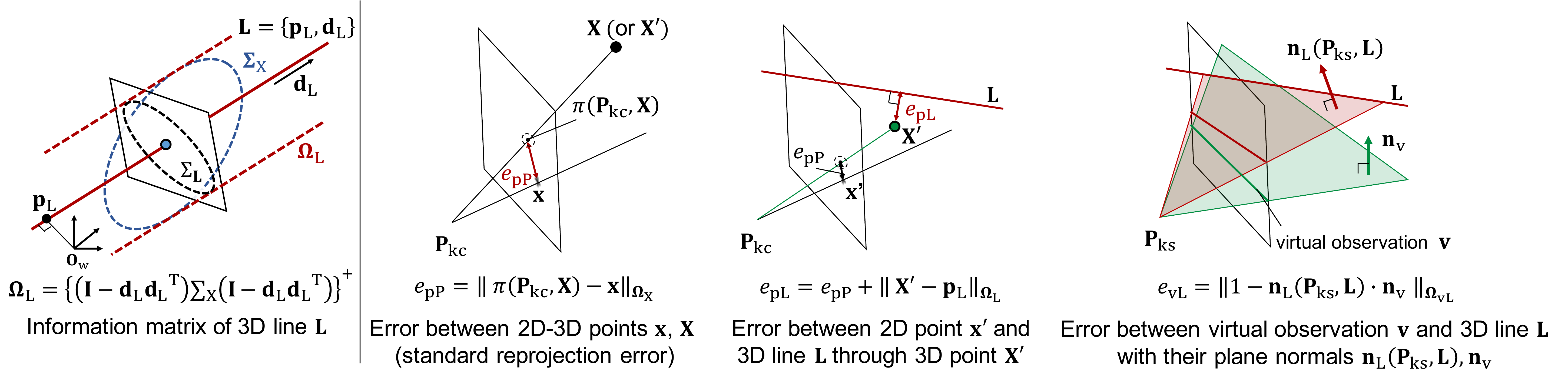} \\
 $\begin{array}{cccc}\scriptsize
 \text{\scriptsize \hspace{12mm}(a) \hspace{10mm}}  & \text{\scriptsize \hspace{12mm}(b) \hspace{10mm}} 
 & \text{\scriptsize \hspace{11mm}(c) \hspace{10mm}} & \text{\scriptsize \hspace{16mm}(d) \hspace{10mm}}\\
 \end{array}$
  \caption{Information matrix of a 3D line and error metrics for bundle adjustments. (a) information matrix of 3D line $\mathbf{L}$, (b) error between 2D-3D points $\mathbf{x}$, $\mathbf{X}$, (c) error between 3D point $\mathbf{x}'$ and 3D line $\mathbf{L}$ through 3D point $\mathbf{X}'$ and (d) error between virtual observation $\mathbf{v}$ and 3D line $\mathbf{L}$ with their plane normals $\mathbf{n}_\text{L}(\mathbf{P}_\mathrm{ks},\mathbf{L})$, $\mathbf{n}_\text{v}$.}
  \label{fig:reprojection_erros}
\end{figure}

\subsubsection{Definition:} 
A prebuilt map contains the 3D lines $\mathbf{L}=\{\mathbf{p}_\mathrm{L},\mathbf{d}_\mathrm{L}\}$. 
They are converted from the 3D points $\mathbf{X}_\mathrm{L}$, whose covariance matrix is defined as $\mathbf{\Sigma}_\mathrm{X_{L}}$. 
The vectors $\mathbf{p}_\mathrm{L},\mathbf{d}_\mathrm{L}$ represent the base point and the directional vector, respectively.
To conceal the information regarding the coordinates in the direction $\mathbf{d}$ of the original 3D points $\mathbf{X}_\mathrm{L}$, we introduce an information matrix of the 3D line $\mathbf{L}$:
\footnotesize
\begin{equation}
\mathbf{\Omega}_\mathrm{L} = \{(\mathbf{I}-\mathbf{d}\mathbf{d}^\mathsf{T})\mathbf{\Sigma}_\mathrm{X_\mathrm{L}}(\mathbf{I}-\mathbf{d}\mathbf{d}^\mathsf{T})\}^\mathsf{+},
\end{equation}
\normalsize
where $\mathbf{A}^+$ is the pseudo-inverse matrix of $\mathbf{A}$ [Fig. \ref{fig:reprojection_erros}(a)].
The information value of $\mathbf{\Omega}_\mathrm{L}$ in the direction $\mathbf{d}$ is zero.

In a standard BA [Fig. \ref{fig:reprojection_erros}(b)], a reprojection error for optimizing camera poses and 3D points is defined as
\footnotesize
\begin{equation}
    e_\mathrm{pP} (\mathbf{P}_\mathrm{kc},\mathbf{X},\mathbf{x}) \coloneqq \| \pi (\mathbf{P_\mathrm{kc},\mathbf{X})-\mathbf{x}} \|^2_\mathbf{\Omega_{\mathrm{X}}},
    \label{eq:standard-reproj_error}
\end{equation}
\normalsize
where $\pi(\cdot)$ is the projection function, $\mathbf{x}$ is the observation points from which the 3D points $\mathbf{X}$ are reconstructed, $\mathbf{\Omega}_{\mathrm{X}}=\mathbf{\Sigma}_{\mathrm{X}}^{-1}$, and $\| \mathbf{e}\|^2_{\mathbf{\Omega}_{\mathrm{X}}}=\mathbf{e}^\mathsf{T}\mathbf{\Omega}_{\mathrm{X}}\mathbf{e}$.

BA with a line cloud, however, requires an error metric between a 2D point $\mathbf{x'}$ and a 3D line $\mathbf{L}$, where $\mathbf{x'}$ is the observation points from which the 3D points $\mathbf{X'}$ are reconstructed.
Hence, as shown in Figure \ref{fig:reprojection_erros}(c), we define the error metric using the 3D point $\mathbf{X'}$, which is initially reconstructed as the intermediate point between the viewing direction of $\mathbf{x'}$ and $\mathbf{L}$, as
\footnotesize
\begin{equation}
    e_\mathrm{pL}(\mathbf{P}_\mathrm{kc},\mathbf{X'},\mathbf{x'},\mathbf{p}_\text{L}) \coloneqq e_\mathrm{pP}(\mathbf{P}_\mathrm{kc},\mathbf{X'},\mathbf{x'}) + \| \mathbf{X'}-\mathbf{p}_\text{L} \|^2_\mathbf{\Omega_{\mathrm{L}}}.
    \label{eq:error_2dp-3dl}
\end{equation}
\normalsize
The first term of Eq. (\ref{eq:error_2dp-3dl}) is the standard reprojection error, which is the constraint between the 2D point $\mathbf{x'}$ and the reconstructed 3D point $\mathbf{X'}$, while the second term is the constraint between the 3D point $\mathbf{X'}$ and the 3D line $\mathbf{L}$.
Through this error metric, the 3D line $\mathbf{L}$ and the camera pose $\mathbf{P}_\mathrm{kc}$ can constrain each other.

Furthermore, the prebuilt line-cloud map may contain errors such as scale drift, which additional observations by other clients can correct. 
However, observation points on the image coordinates should be dropped when a user uploads a line cloud to a server because they can recover the corresponding 3D points.
As a result, there is no constraint between the 3D lines $\mathbf{L}$ and their keyframe camera poses $\mathbf{P}_\mathrm{ks}$ for a BA.

Here, we introduce the virtual observation $\mathbf{v}$, which is the projection of the initial 3D line $\mathbf{L}_\mathrm{0}$ onto the keyframe.
Strictly speaking, $\mathbf{v}$ is represented by the normal vector $\mathbf{n}_\text{v}$ of the plane defined by the camera center and $\mathbf{L}_\mathrm{0}$.
Similarly, for the current state, the normal vector of the plane defined by the camera center and the 3D line $\mathbf{L}$ is represented as $\mathbf{n}_\text{L}(\mathbf{P}_\mathrm{ks},\mathbf{L})$.
We define the error metric between the 3D lines $\mathbf{L}$ and the virtual observation $\mathbf{v}$ with their normals as
\footnotesize
\begin{equation}
    e_\mathrm{vL}  (\mathbf{P}_\mathrm{ks},\mathbf{L},\mathbf{v}) \coloneqq \| 1 - \mathbf{n}_\text{L} (\mathbf{P}_\mathrm{ks},\mathbf{L}) \cdot \mathbf{n}_\text{v} \|^2_\mathrm{\Omega_{\mathrm{vL}}}.
    \label{eq:error_virtual}
\end{equation}
\normalsize
It should be noted that Eq. (\ref{eq:error_virtual}) is directly applicable to other projection models, such as the equirectangular model, because it is defined in the local camera coordinates.

\begin{figure}[!t]
\hfil
$\begin{array}{ccc}\scriptsize
  \includegraphics[width=2.5cm]{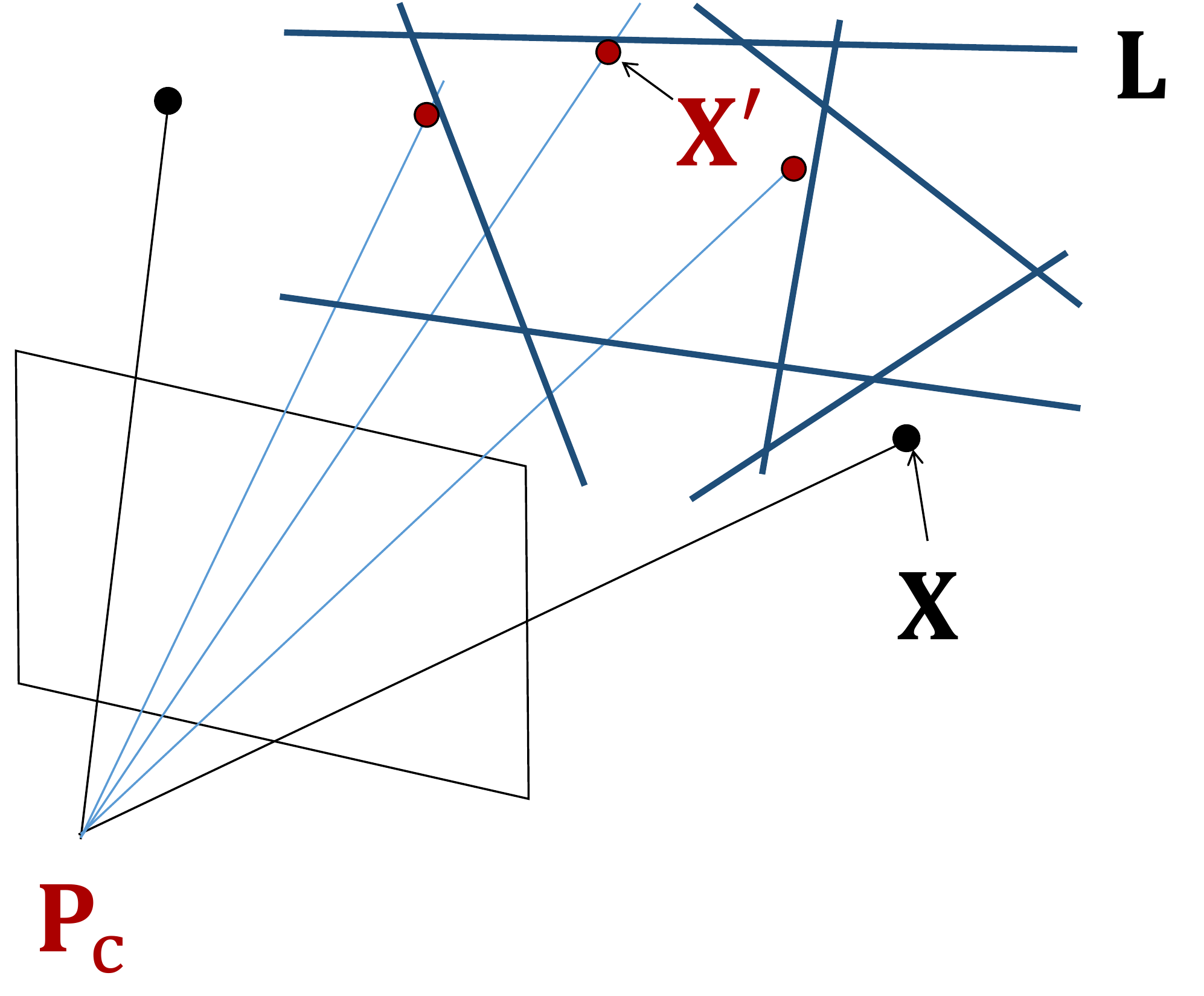} 
&
\hspace{4mm}  \includegraphics[width=3.0cm]{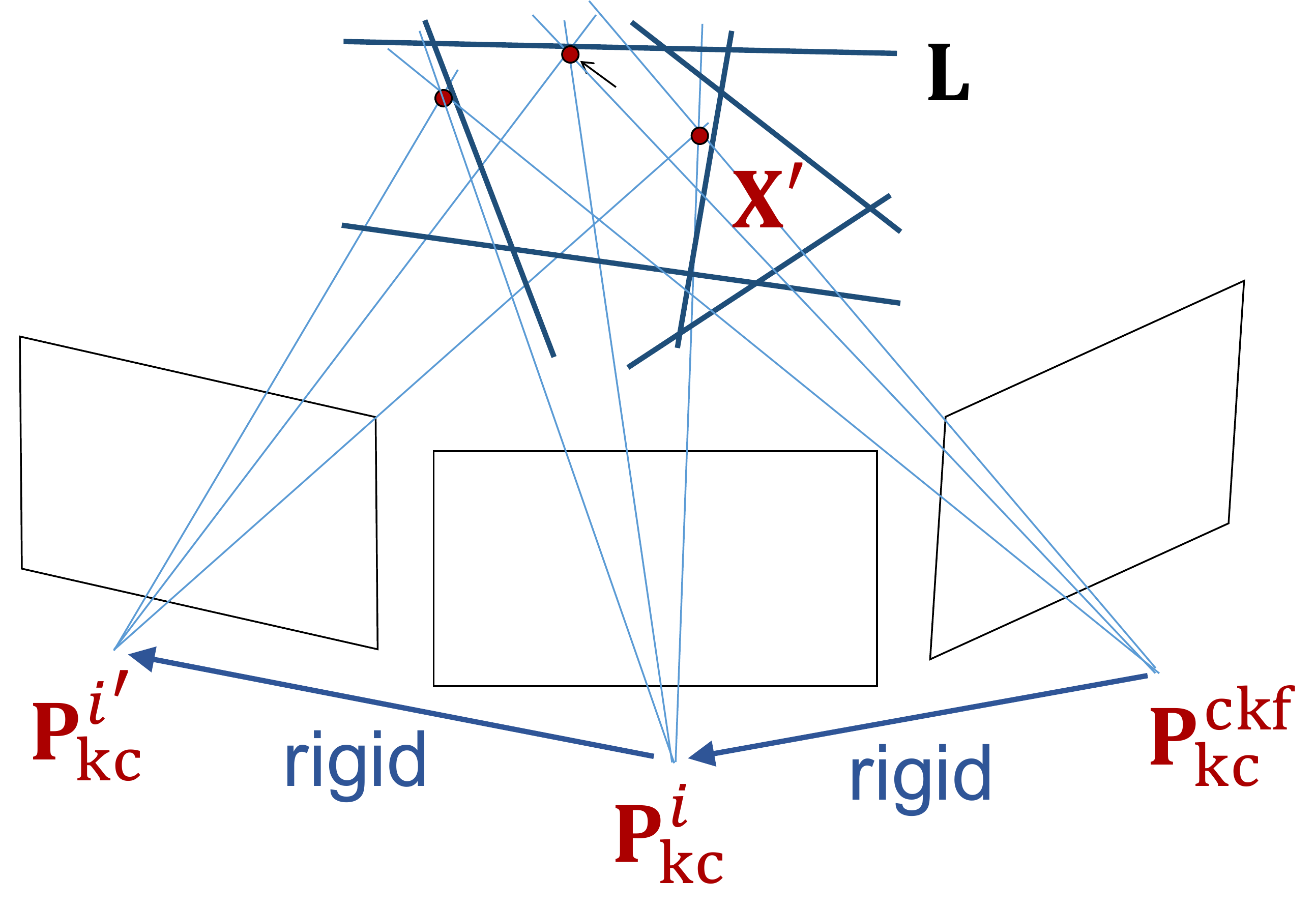} \hspace{2mm}
&
  \hspace{2mm} \includegraphics[width=3.0cm]{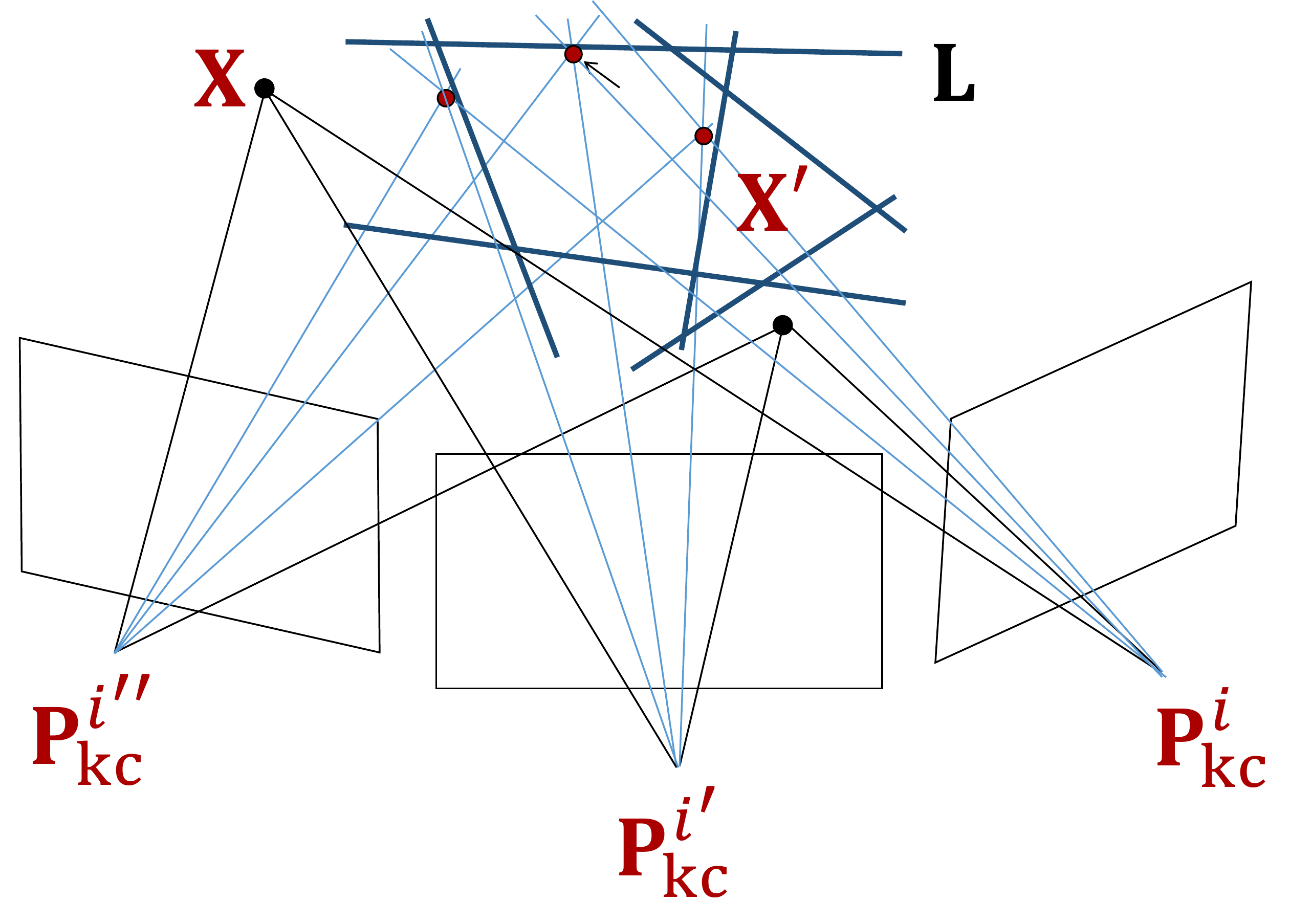}\\
\text{\scriptsize \hspace{0mm}(a) Motion-only BA \hspace{0mm}}  & \text{\scriptsize \hspace{10mm}(b) Rigid-stereo BA\hspace{7mm}} & \text{\scriptsize \hspace{7mm}(c) Local BA\hspace{8mm}} \\
\end{array}$ \\
\vspace{2mm}\\
$\begin{array}{cc}\scriptsize
  \includegraphics[width=8.0cm,bb=0 0 1737 552]{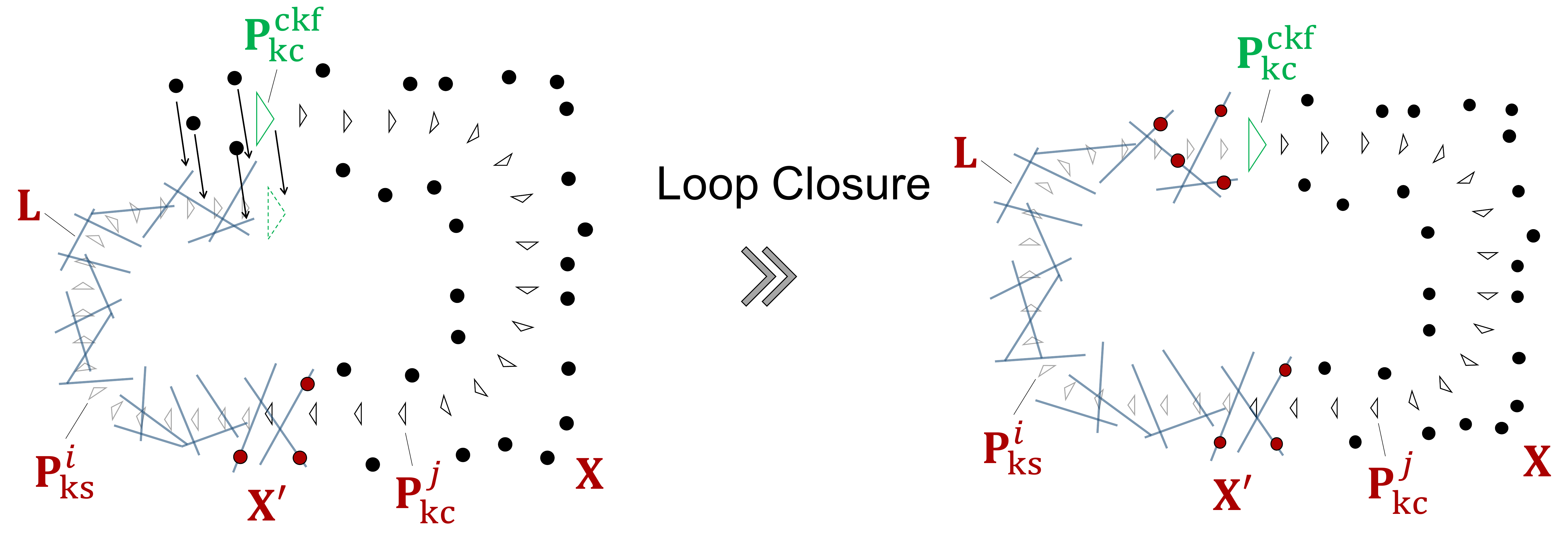}
&
\includegraphics[width=4.0cm,bb=0 -50 678 292]{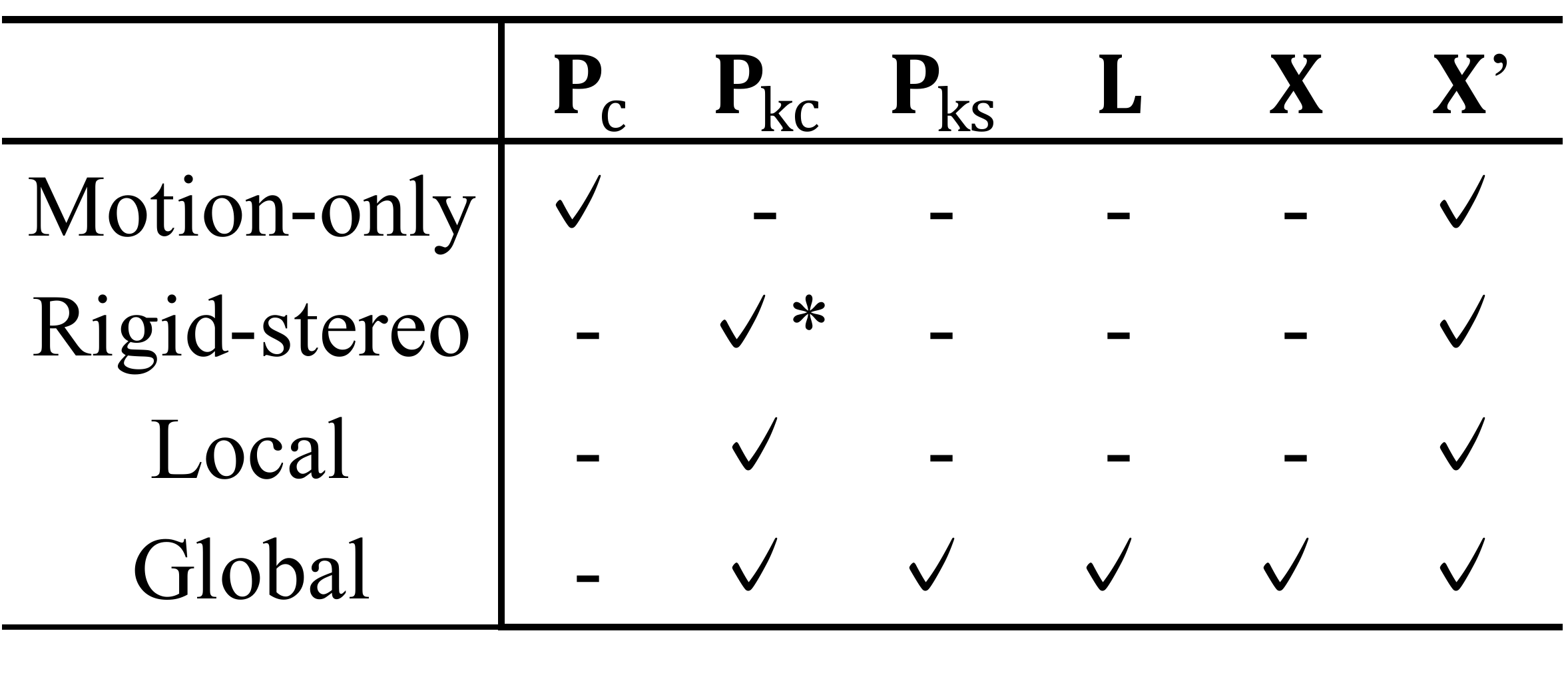} \\
\text{\scriptsize \hspace{0mm}(d) Global BA before and after loop closure \hspace{0mm}} 
& 
\text{\scriptsize \hspace{0mm} Optimization parameters \hspace{0mm}} 
\\
\end{array}$

  \caption{Optimization parameters of BAs for: (a) motion-only, (b) rigid-stereo, (c) local, and (d) global (in red). * Only the rigid-stereo BA is set under the rigid constraint on the relative camera poses between keyframes.}
  \label{fig:all-ba}
\end{figure}

Figure \ref{fig:all-ba} shows the optimization parameters in each BA with a line cloud: (a) Motion-only BA, (b) Rigid-stereo BA, (c) Local BA and (d) Global BA.
LC-VSLAM, as the standard Visual SLAM, reconstructs 3D points as $\mathbf{X}$, which are corresponded with 3D lines $\mathbf{L}$ in relocalization and loop detection. 
The corresponded 3D points $\mathbf{X}$ are identified as $\mathbf{X'}$ for the optimization.
Utilizing the above error metrics, the four bundle adjustments with a line cloud are defined as follows.

\subsubsection{Motion-only BA:}
\label{subsubsec:motion-only-ba}
A tracking thread estimates the camera pose for each input frame. 
For real-time tracking, as shown in Figure \ref{fig:all-ba}(a), the motion-only BA optimizes only the camera pose of the current frame $\mathbf{P}_\mathrm{c}$ and the reconstructed 3D points $\mathbf{X}'$ with fixed 3D points $\mathbf{X}$ and lines $\mathbf{L}$ as
\footnotesize
\begin{equation}
\mathbf{P}_\mathrm{c}^*, \mathbf{X'}^* = \argmin_{\mathbf{P}_\mathrm{c}^*, \mathbf{X'}^*} 
\sum_{j}
e_\mathrm{pP} (\mathbf{P}_\mathrm{c},\mathbf{X}^j,\mathbf{x}^j) +
\sum_{k} e_\mathrm{pL}(\mathbf{P}_\mathrm{c},\mathbf{X'}^k,\mathbf{x'}^k,\mathbf{p}_\text{L}^k),
\label{eq:motion-only-ba}
\end{equation}
\normalsize
where $j,k$ indicate the indices of the 3D points $\mathbf{X}, \mathbf{X'}$ which are visible from the current frame, respectively.
The first term pertains to the constraints between the camera pose of the current keyframe $\mathbf{P}_\mathrm{c}$ and the 3D points $\mathbf{X}'$ that have been already reconstructed with the previous frames. The second one refers to those between the camera pose $\mathbf{P}_\mathrm{c}$ and the 3D lines $\mathbf{L}$.

\subsubsection{Rigid-stereo BA:}
\label{subsubsec:stereo-ba}
For relocalization and loop detection, the local 3D points $\mathbf{X}$, which are merged as $\mathbf{X'}$ after matching with the 3D lines $\mathbf{L}$, have already been reconstructed and locally optimized with the local keyframes.
As shown in Figure \ref{fig:all-ba} (b), the rigid-stereo BA can optimize the camera pose of the current keyframe $\mathbf{P}_\mathrm{kc}^\mathrm{ckf}$ and the 3D points $\mathbf{X'}$ as
\footnotesize
\begin{equation}
\mathbf{P}_\mathrm{kc}^{\mathrm{ckf}*}, \mathbf{X'}^* = \argmin_{\mathbf{P}_\mathrm{kc}^\mathrm{ckf}, \mathbf{X'}} 
\sum_{i}\sum_{j} e_\mathrm{pL}(\Delta \mathbf{P}^i \mathbf{P}_\mathrm{kc}^\mathrm{ckf},\mathbf{X'}^{j},\mathbf{x'}^{i,j},\mathbf{p}_\text{L}^j)
\label{eq:stereo-ba}
\end{equation}
\normalsize
where $\Delta \mathbf{P}^i = \mathrm{const.}$ is the relative camera pose between the current keyframe and the $i$-th neighboring keyframe which shares the 3D points ($\Delta \mathbf{P}^i=\mathbf{I}$ if $i=\mathrm{ckf}$).

\subsubsection{Local BA:}
\label{subsubsec:local-ba}
The rigid-stereo BA is a special case of the local BA.
In the local mapping thread, new keyframes of the client $\mathbf{P}_\mathrm{kc}$ are inserted, and the 3D points $\mathbf{X}$ are newly reconstructed from only their 2D points.
Hence, as shown in Figure \ref{fig:all-ba} (c), the camera pose of the local keyframes $\mathbf{P}_\mathrm{kc}$ and their 3D points $\mathbf{X}$ and $\mathbf{X'}$ are optimized as
\footnotesize
\begin{align}
\mathbf{P}_\mathrm{kc}^*, \mathbf{X}^*, \mathbf{X'}^* =
& \argmin_{\mathbf{P}_\mathrm{kc},\mathbf{X},\mathbf{X'}} 
\sum_{i}\sum_{j} e_\mathrm{pP}(\mathbf{P}_\mathrm{kc}^i,\mathbf{X}^{j},\mathbf{x}^{i,j}) \nonumber \\ 
 + & \sum_{i}\sum_{k} e_\mathrm{pL}(\mathbf{P}_\mathrm{kc}^i,\mathbf{X'}^{k},\mathbf{x'}^{i,k},\mathbf{p}_\text{L}^k).
 \label{eq:local-ba}
\end{align}
\normalsize

\subsubsection{Global BA:}
\label{subsubsec:global-ba}
After loop detection and pose graph optimization \cite{grisetti2010pgo,strasdat2010scale}, the camera poses and 3D structures, which include the 3D lines $\mathbf{L}$ and the camera poses of their keyframes $\mathbf{P}_\mathrm{ks}$, are globally optimized with the virtual observation $\mathbf{v}$ and its error metric $e_\mathrm{vL}$ as
\footnotesize
\begin{align}
\mathbf{P}_\mathrm{kc}^*, \mathbf{P}_\mathrm{ks}^*, \mathbf{X}^*, \mathbf{X'}^*, \mathbf{L}^* =
& \argmin_{\mathbf{P}_\mathrm{kc}, \mathbf{P}_\mathrm{ks},\mathbf{X},\mathbf{X'}, \mathbf{L}} 
\sum_{i}\sum_{j} e_\mathrm{pP}(\mathbf{P}_\mathrm{kc}^i,\mathbf{X}^{j},\mathbf{x}^{i,j}) \nonumber \\
& \hspace{-20mm} +  \sum_{i}\sum_{k} e_\mathrm{pL}(\mathbf{P}_\mathrm{kc}^i,\mathbf{X'}^{k},\mathbf{x'}^{i,k},\mathbf{p}_\text{L}^{k}) + \sum_{i'}\sum_{l} e_\mathrm{vL}  (\mathbf{P}_\mathrm{ks}^{i'},\mathbf{L}^{l},\mathbf{v}^{i',l}),
\label{eq:global-ba}
\end{align}
\normalsize
where $i, i'$ are the indices of all client and server keframes, respectively, and $l$ is the index of the 3D line $\mathbf{L}$.

\section{Experiments}
\label{sec:experiments}

\subsection{Experimental Setting}

The performance of LC-VSLAM was tested to quantitatively and qualitatively evaluate from multiple perspectives (see the algorithm in Sec.~\ref{sec:proposed_method}). We have carried out all experiments with a Core i9-9900K (8 cores @ 3.60GHz) with a 64 GB RAM.
Considering its practical usability, we evaluated the performance from the following viewpoints.

\subsubsection{Tracking time:}
We evaluated the tracking time of each frame to confirm the real-time performance of the proposed framework.
Based on the mean tracking time, we compared LC-VSLAM with p6L, which applies a single-shot localization algorithm in the 3D line cloud to every frame~\cite{speciale2019privacylocalization}.

\subsubsection{Accuracy of camera pose estimation:}
We evaluated the accuracy of camera poses after a local map was registered to a prebuilt map for LC-VSLAM and the previous method.
First, a local 3D point-cloud map was created by a client, and geometrically registered to a global 3D line-cloud map downloaded from the server using the estimated 3D transformation between them.
After the registration, the camera pose accuracy was evaluated using the latest keyframe of the registered local map in the coordinates system of the global map.

\subsubsection{Comparison to a conventional Visual SLAM system:}
The foregoing perspectives were evaluated run on the synthetic dataset generated by the CARLA Simulator~\cite{dosovitskiy2017carla} and the real image dataset KITTI \cite{geiger2012kitti}.
We also compared three camera types (perspective, fisheye, and equirectangular) in the evaluation to confirm that the proposed algorithms work well for various types of projection models.

\subsection{Implementation Details}
We implemented the LC-VSLAM system based on OpenVSLAM \cite{openvslam2019} by integrating the four dedicated modules as follows:
(i) add the data structure of a line, such as line direction and covariance,
(ii) add the P4L solver and the rigid-stereo BA [Eq. (\ref{eq:stereo-ba})] to the modules of the loop detector and the relocalizer,
(iii) adapt the one-to-many feature matching to the many-to-many one, and
(iv) replace the cost functions in the motion-only, local, and global BAs with those of [Eqs. (\ref{eq:motion-only-ba}), (\ref{eq:local-ba}), and (\ref{eq:global-ba})].

\subsection{Dataset and Prebulit Map Creation}
All the evaluation was performed on our new CARLA dataset because there are no publicly available benchmarks for evaluating LC-VSLAM. 
The dataset should satisfy the following two requirements to evaluate the effectiveness of LC-VSLAM:
(i) a sequence pair contains sufficient overlaps and loops to allocate a sequence for a prebuilt map and the other for an input of LC-VSLAM to evaluate the tracking and the loop closure, and
(ii) image sequences of various types of camera models are available to evaluate the versatility on projection models.
KITTI camera stereo dataset \cite{geiger2012kitti} is one of the publicly available datasets for evaluating accuracy of Visual SLAM systems and meets the requirement (i). 
However, in the dataset, the baseline is very short, and the image pairs are synchronized, which make tracking too easy. 
The KITTI dataset contains only perspective projection images, and thus does not meet the requirement (ii).
Therefore, we performed all quantitative evaluations on our new CARLA datataset, 
and verified the LC-VSLAM's applicability to real image datatasets on the KITTI stereo dataset.
Our {\it Desk} and {\it Campus} datasets were used to qualitatively evaluate the effectiveness of the LC-VSLAM on real scenes. 

\subsubsection{CARLA Dataset:}
We used the CARLA simulator \cite{dosovitskiy2017carla} to create a dataset for evaluating the accuracy of our tracking and bundle adjustment methods, which utilize a line cloud as a prebuilt map (see the supplementary material for details).
The simulator allows synthesis of photo-realistic images with the camera poses of outdoor scenes.
Hence, to evaluate the tracking time, we generated an image sequence pair ($\#$01) which almost overlaps each other with small displacements.
Additionally, we created eight pairs of mid-scale image sequences ($\#$02-09) for each camera type to evaluate the localization accuracy and created three pairs of large-scale image sequences with loop-closure points ($\#$10-12) to evaluate the effectiveness of the global bundle adjustment.
The sequence pairs ($\#$02-12) satisfy the predifined requirements and each pair partially overlaps each other, exclusive of the $\#01$ pair because there is no loop-closure point between the sequences.
This dataset will be publicly available.

\subsubsection{KITTI Dataset}
To evaluate the effectiveness of LC-VSLAM, we selected two sequences of the KITTI stereo dataset also meeting the requirement (i), \#00 and \#05.
We prebuilt maps with the odd-numbered images of the left camera and input the even-numbered images of the right camera to LC-VSLAM.
The prebuilt maps were constructed as follows: (I) perform a standard Visual SLAM to estimate initial camera poses and 3D points,
(II) replace the estimated camera poses with the ground truth, 
(III) perform a bundle adjustment to correct the errors found in the ground truth and to refine the positions of the 3D points.

\subsubsection{Campus and Desk Datasets}
We also created two datasets for qualitative evaluations.
The sequences of Campus dataset (Scene A and B) were captured by cameras with wide-angle and fisheye lenses (Panasonic LUMIX GX7MK3) and a
panoramic camera (RICOH THETA Z1) independently.
Their common camera path included both indoor and outdoor scenes.
The sequences of Desk dataset were captured by a camera with a wide angle lens, and a pair of successive sequences was processed to assure the privacy protection by means of removing two personal objects in the original image and changing the displays.

\begin{table}[t]
  \centering
  {\scriptsize
  \caption{Tracking time of each frame, mean absolute pose errors (APE) for translation [m] and rotation [deg] of the single-shot localization by p6L~\cite{speciale2019privacylocalization} and LC-VSLAM. The image resolution is $640\times360$. }
    \begin{threeparttable}[h]
    \begin{tabular}{C{25mm} " C{30mm} C{45mm}}
      \thickhline
        & Tracking time [ms] & APE for trans. [m] / rot. [deg] \\\thickhline
      \hspace{1mm} p6L \cite{speciale2019privacylocalization} & 140.3  & 0.7815 / 0.5896 \\
      \hspace{1mm} LC-VSLAM (ours) & \textbf{31.09} & \textbf{0.1979} / \textbf{0.2841} \\\thickhline
    \end{tabular}
        \label{table:tracking_time_accuracy}
    \end{threeparttable}
  }
    \vspace{0mm}
\end{table}


  \begin{table}[t]
  \centering
  {\scriptsize
  \caption{Mean absolute pose errors (APE) for translation [m] and rotation [deg] of synthetic images by the CARLA simulator and KITTI dataset. Lower is better.}
    \begin{threeparttable}[h]
    \begin{tabular}{L{20mm}" C{20mm} C{20mm} C{20mm}" C{20mm}}
      \thickhline
      \hspace{1mm} & & CARLA & & KITTI \\
      \thickhline
      \hspace{1mm} Prebuilt map & Perspective & Fisheye & Equirectangular & Perspective \\
      \thickhline
      \hspace{1mm} 3D points & 3.290 / 0.6273  & 2.883 / 0.4402 & 3.079 / 0.2375 & 3.801 / 1.012 \\
      \hspace{1mm} 3D lines (ours) & 3.651 / 0.8416 & 3.177 / 0.5941 & 3.075 / 0.2766 & 4.488 / 1.309  \\
      \thickhline
    \end{tabular}
        \label{table:ape}
    \end{threeparttable}
  }
    \vspace{-2mm}
  \end{table}

\subsection{Quantitative Evaluation}
We quantitatively evaluated the tracking time of each frame and means of the absolute pose errors (APE) for translation and rotation of the single-shot localization by the proposed LC-VSLAM and p6L~\cite{speciale2019privacylocalization} on the sequence pair $\#01$ (Table \ref{table:tracking_time_accuracy}). 
In $640\times360$ image resolution, the tracking time for LC-VSLAM is 31.09 ms ($\approx$32[fps]), much faster than the 140.3 ms for p6L, which can be defined as {\it real time}.

LC-VSLAM also achieves a better result in APE, 0.1979/0.2841 [m]/[deg], than p6L does: 0.7815/0.5896 [m]/[deg].
This means the proposed method outperforms p6L in its tracking speed and localization accuracy because LC-VSLAM can utilize the continuity of input images.
Moreover, Table \ref{table:ape} shows the APE for translation and rotation on synthetic images via the CARLA simulator 
for each camera device ($\#$02-09) and on real images of the KITTI stereo datatset for perspective cameras.
LC-VSLAM can estimate camera poses using a 3D line map with accuracy similar to using a 3D point map for all the camera types. 
In the case of fisheye projection, the estimation error is relatively larger than that of other projection models. 
These results verify the accuracy and the efficiency of the LC-VSLAM.

To evaluate the effectiveness of the global bundle adjustment with a line cloud, we compared the localization accuracy of LC-VSLAM with and without the pose graph optimization (PGO) and the global bundle adjustment on the sequence pairs of the CARLA dataset ($\#$10-12).
Table \ref{table:loop_closure_ape_trans} shows the mean APE for translation and rotation on three sequences.
The PGO corrects the estimation errors, especially for the APE, as with a standard Visual SLAM, and our global BA can refine the PGO results.

\begin{table}[t]
  \centering
   \caption{Mean APE for trans. [m] and rot. [deg] of LC-VSLAM with/without the pose graph optimization (PGO) and the global bundle adjustment (Global BA) for each camera device data.}
       \label{table:loop_closure_ape_trans}
  {\scriptsize
     \begin{threeparttable}[h]
   \begin{tabular}{c"c|c|c} 
     \thickhline
      & Perspective & Fisheye & Equirectangular
      \\\thickhline
  None  & 24.06 / 1.292  & 10.16 / 1.064 & 14.28 / 3.682 \\\hline
  Only PGO & 3.301 / 1.151  & 1.670 / \textbf{0.8039} & 9.640 / 2.790 \\\hline
  PGO \& Global BA & \textbf{3.018} /  \textbf{1.100} & \textbf{1.593} / 0.8525 & \textbf{8.320} / \textbf{2.404} \\\thickhline
   \end{tabular}
   \end{threeparttable}
  }
   \end{table}

\subsection{Qualitative Evaluation}
\begin{figure*}[t]
    \centering
    \includegraphics[width=\linewidth]{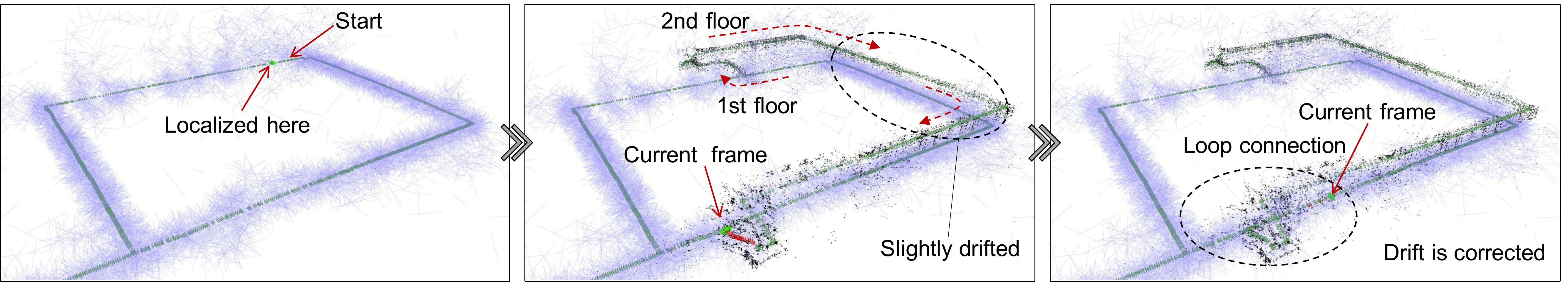} \\
    $\begin{array}{ C{40mm} C{40mm} C{40mm}}\scriptsize
     \text{\hspace{-1mm}\scriptsize (a) Relocalization}
    &
    \text{\hspace{-3mm}\scriptsize (b) Before loop closure}
    &
    \text{\hspace{-4mm} \scriptsize (c) After loop closure} \\
    \end{array}$
    \caption{Example of a reconstructed 3D map in case of a equirectangular model, which includes a prebuilt 3D line cloud (blue), a reconstructed 3D point cloud (black), and keyframes (green or red). (See all the other results in the supplementary material.)}
    \label{fig:tit_build_three_seqs}
    \vspace{-10pt}
\end{figure*}

We applied the proposed framework to various scenes, and confirmed that the algorithms work effectively.
The images of Figure~\ref{fig:tit_build_three_seqs} show the scenes with a prebuilt 3D line cloud (blue), a reconstructed 3D point cloud (black),
and keyframes (green or red), which were all made from the video sequences captured with the panoramic camera (Scene A of Campus dataset).
The order from the first (left) to the last (right) columns represents an example of the reconstruction process.

LC-VSLAM localizes and tracks the camera in the prebuilt 3D line-cloud map soon after each sequence starts.
Additionally, mapping as well as tracking continuously perform well even in the area outside of the prebuilt map.
In other words, the prebuilt map can be extended effectively with the LC-VSLAM processing on the client.
Finally, a loop connection point is correctly found when the camera goes back to the prebuilt map area.

Furthermore, in the tracking process of our framework, 3D points may be subsequently restored near 3D lines in a prebuilt map.
In a case wherein an object to be protected against inversion attack is present in the prebuilt map but does not in a sequence that a client captures, it is necessary to guarantee that the 3D points of the concealed privacy objects are not restored near the corresponding 3D lines.
Figure \ref{fig:abst_lcvslam} shows a typical example of the case with the Desk dataset and that privacy related to an object that only exists in the prebuilt map is protected.

The foregoing results lead us to believe that the proposed LC-VSLAM framework works well for various scenes and cameras.

\section{Conclusions}
\label{sec:conclusion}
In this paper, we proposed a privacy-preserving Visual SLAM framework for real-time tracking and bundle adjustment with a line-cloud map, which we refer to as LC-VSLAM. 
In the framework, we have presented efficient methods of relocalization and tracking by utilizing 3D points reconstructed by a Visual SLAM client and discretizing 3D lines to 3D points.
For optimization in terms of both 3D points and lines, we proposed four types of bundle adjustments by introducing error metrics for 3D lines.
These methods are applicable to various types of projection models, such as perspective and equirectangular models.
The experiments on videos captured with various types of cameras verified the effectiveness and the real-time performance of LC-VSLAM.
Thus, the proposed framework enables real-time tracking/mapping with a line-cloud map in practical applications such as AR and MR.
The protective function of scene privacy is in place for map sharing among multiple users.

For future studies, we will refine the formulation for the error metric of the virtual observation $\mathbf{v}$.
In this study, $\mathrm{\Omega}_\mathrm{vL}$ was set as a constant value because the methodology is not trivial to convert the information matrix of the 3D line $\mathbf{L}$ to that of the cosine distance between their plane normals.
The refined formulation will enable a more accurate global optimization with prebuilt line clouds.

\clearpage
\bibliographystyle{splncs04}
\bibliography{egbib}

\end{document}